\pdfoutput=1

\documentclass[11pt]{article}

\usepackage{EMNLP2022}

\usepackage{times}
\usepackage{latexsym}
\usepackage{multirow}
\usepackage{booktabs}

\usepackage[T1]{fontenc}

\usepackage[utf8]{inputenc}

\usepackage{microtype}

\usepackage{inconsolata}

\usepackage{adjustbox}

\title{Findings of the TSAR-2022 Shared Task\\ on Multilingual Lexical Simplification}

\author{Horacio Saggion$^1$, Sanja \v{S}tajner$^2$, Daniel Ferr\'{e}s$^1$, Kim Cheng Sheang$^1$\\
{\bf Matthew Shardlow}$^3$,
{\bf Kai North}$^4$, {\bf Marcos Zampieri}$^4$\\
$^1$Universitat Pompeu Fabra, Barcelona, Spain\\
$^2$Karlsruhe, Germany\\
$^3$Manchester Metropolitan University, Manchester, UK\\
$^4$George Mason University, Fairfax, VA, USA\\
  \texttt{horacio.saggion@upf.edu} \\}

\begin{document}
\maketitle
\begin{abstract}
We report findings of the TSAR-2022 shared task on multilingual lexical simplification, organized as part of the Workshop on Text Simplification, Accessibility, and Readability TSAR-2022 held in conjunction with EMNLP 2022. The task called  the Natural Language Processing research community to contribute with methods to advance the state of the art in multilingual lexical simplification for English, Portuguese, and Spanish. 
A total of 14 teams submitted the results of their lexical simplification systems for the provided test data. Results of the shared task indicate new benchmarks in Lexical Simplification with English lexical simplification quantitative results noticeably higher than those obtained for Spanish and (Brazilian) Portuguese.
\end{abstract}

\section{Introduction}
 Lexical Simplification \cite{shardlowsurvey2014,paetzoldsurvey2017} is a sub-task of Automatic Text Simplification \cite{Saggion'2017} that aims at replacing difficult words with easier to read (or understand) synonyms while preserving the information and meaning of the original text.  This is a key task to facilitate reading comprehension to different target readerships such as foreign language learners, native speakers with low literacy levels or people with different  reading impairments (e.g. dyslexic individuals). As such, it has gained considerable attention in the past few years \citep{stajner-2021-automatic}.
 
Although Lexical Simplification systems  can be developed following different architectural precepts, several studies  have  suggested the following pipe-lined approach:  
\begin{enumerate}
\item identification of complex terms (Complex Word Identification - CWI),
\item generation of substitution words (Substitute Generation - SG), 
\item selection of the substitutes that can fit in the context (Substitute Selection - SS),
\item ranking substitutes by their simplicity (Substitute Ranking - SR), and 
\item morphological generation and context adaptation (e.g. agreement).
\end{enumerate}

There exists a considerable body of research in lexical simplification for English 
\citep{horn2014learning,Glavas&Stajner'2015,paetzold-specia-2017-lexical,qiang2020BERTLS}. 
However, and in spite of several lexical simplification studies for languages other than English notably  
\citep{Bott&al'2012,Baeza-YatesRD15,ferresSG_BGNLP2017,ferres-saggion:2022:LREC} for Spanish,  
\citep{Hartmann&al'2018,north2022alexsis} for Portuguese,  
\citep{hmida-etal-2018-assisted} for French, 
\citep{ChineseLS-2021} for Chinese, 
\citep{kajiwara-yamamoto-2015-evaluation,hading-etal-2016-japanese} for Japanese and 
\citep{abrahamsson-etal-2014-medical} for Swedish, 
there is a clear need to broaden the scope of lexical simplification in terms of language coverage.  Moreover, given its social relevance in making information accessible to broader audiences, we believe it is important to understand how far automatic systems can go in this task.

We therefore established this first Shared Task on Multilingual Lexical Simplification calling the NLP research community to contribute with methods to advance the state of the art. 
The task called for systems able to simplify words in context in (one or more of)  three languages, namely English, Portuguese, and Spanish. Systems have to deal with steps 2-5 above to generate, select,  rank, and adapt to context  substitutes for a given complex word in a sentence.  As the result of our call for systems, of the 22 teams registered to the task, 14 sent their system outputs for evaluation. There were 31 different runs for English, 15 for Spanish, and 14 for Portuguese.  

This paper overviews the first Shared Task on Multilingual Lexical Simplification. We describe in detail the task, the trial and test data used, the evaluation metrics, and the results. 
We also provide an analysis of the results and consider possible ways to expand the current scope of the task. 

\subsection{State-of-the-Art Lexical Simplification}

In recent years, researchers have turned to large off-the-shelf word embedding models, instead of pre-compiled lists of synonyms or lexical databases,  for retrieving (or generating) substitution candidates \cite{Glavas&Stajner'2015, Paetzold&Specia'2016a}, ranking them for simplicity and context using  several sorting factors such as frequency, target context similarity, language model probabilities, etc.  These approaches demonstrated better coverage than previous systems. Before the  TSAR 2022 Shared Task,  the state of the art for English lexical simplification was the LSBert system \citep{qiang2020BERTLS}, which used the pre-trained transformer  language model BERT \citep{Devlin&al'2019} and a masking technique for finding suitable simplifications for complex words, resorting, as previous approaches,  to unsupervised ranking using several feature combinations.

Lexical simplification in languages other than English attracted less attention, however several systems for Spanish have been proposed since the 
initial work of \citep{Bott&al'2012}. As it is with  the case of English, here the use of neural systems is also observed. For example, \citet{Alarcon&al'2021} leverages pretrained word embedding vectors and BERT models. Subsystems were developed for  CWI, SG, and SS; in particular, the CWI sub-task was evaluated using the CWI 2018 shared task dataset for Spanish \citep{yimam-etal-2018-report} where it was found that traditional algorithms (i.e. Support vector Machines) are still competitive in this task. 
The SG and SS sub-tasks were evaluated using a portion of 500 instances of the EASIER corpus \citep{alarconIEEEAccess2021}. Each instance of this portion contains a sentence, a target word and three substitutions.
More recently,  \citet{ferres-saggion:2022:LREC} presented ALEXSIS, a dataset for benchmarking Lexical Simplification in Spanish, and performed experiments with several neural and unsupervised systems for the different phases of the simplification pipeline. They also performed the first evaluation of an adaptation of LSBERT \cite{qiang2020BERTLS} software for Spanish for SG and Full Pipeline with the ALEXSIS and EASIER datasets, achieveing state of the art. 

For Brazilian Portuguese, a data-driven machine translation approach has been proposed in \cite{Specia'2010}.
In the current neural paradigm,  \citet{north2022alexsis} developed and evaluated, on a new corpus for Portuguese based on ALEXSIS \citep{ferres-saggion:2022:LREC},  four transformer
models for substitute generation following the BERT masked approach \cite{qiang2020BERTLS}.  Somehow related is the work of \citep{Hartmann&al'2020} that describe a Portuguese 
datasets which is  designed for simplification of  texts for children. 
\subsection{Previous Lexical Simplification Shared Tasks}

The first shared task in lexical simplification was proposed for SemEval 2012. It addressed English Lexical Simplification  \cite{Specia&al'2012} and offered the opportunity to evaluate systems able to rank
substitution candidates in relation to their simplicity. It was, therefore, concentrating just on step number 4 in the lexical simplification pipeline we have described in the Introduction.  The dataset used was taken from the Lexical Substitution task at SemEval 2007 which was enriched with simplicity rankings provided by second language learners with high proficiency levels in English, rankings per instance were aggregated to obtain a final gold annotation. The task attracted 5 different institutions which provided nine systems in total.


Complex word identification (CWI), which is not addressed in the current TSAR challenge,
has been explored in two shared tasks: SemEval 2016 CWI for English \citep{Paetzold&Specia'2016a}, and the BEA 2018 CWI shared task for multiple languages \citep{yimam-etal-2018-report}. 
In SemEval 2016 CWI task, participants were requested to predict which words in a given sentence would be considered complex by a non-native English speaker. A CWI dataset composed of 9,200 instances was created with sentences from different datasets which have already been used in text simplification research and it  was, for the task objectives,  annotated by non-native speakers of English. The task attracted 21 teams which produced a total of 42 systems.  The BEA 2018 CWI shared task proposed to tackle CWI in English, German, and Spanish (training and test data were provided),  together with a multilingual task with French as a target language without training data. Teams were asked to produce systems to classify words as either complex or simple (binary) and/or provide a probability for the complexity  of each word. The shared task attracted 11 teams. 
The SemEval 2021 shared task on lexical complexity prediction \citep{shardlow-etal-2021-semeval} also provided a new dataset for complexity detection for single words and multi-word expressions in English attracting 55 teams. Additionally, the IberLef 2020 
forum proposed a shared task on Spanish complex word identification \cite{Zambrano&Montejo'2020} but attracted few participants. 

\section{Task Description}

The TSAR-2022 shared task featured three tracks:

\begin{itemize}
    \item Lexical simplification in English;
    \item Lexical simplification in Spanish; 
    \item Lexical simplification in (Brazilian) Portuguese.
\end{itemize}

In all tracks, the task was the same. Given a sentence/context and one target (complex) word in it, provide substitutes for the target word that would make the sentence easier to understand. It was allowed to submit up to 10 substitutes, ordered from the best to the least fitting/simple one.
Ties were not allowed.

Participants were provided with several trial examples in each language. Training datasets were not provided. However, participants were allowed to use any external resources for building their lexical simplification systems. Participating systems were evaluated on test sets using several metrics.\footnote{Compilation of the datasets used in the TSAR-2022 shared task, their limitations, and strong baselines for English, Spanish, and (Brazilian) Portuguese are described in detail in \citep{StajnerEtAl-2022-FrontiersAI-LS}.}

\subsection{Datasets}
Datasets for all three languages were compiled using comparable procedures. 

\subsubsection{Context and Target Word Selection}
For English and Spanish, the sentences/contexts and target words were selected from respective datasets used in the BEA-2018 shared task on complex word identification \citep{yimam-etal-2018-report}.\footnote{https://sites.google.com/view/cwisharedtask2018} For (Brazilian) Portuguese, the sentences and target words were selected from the PorSimplesSent dataset \citep{leal-etal-2018-nontrivial}. In the (English and Spanish) CWI-2018 datasets, complex words were marked based on a crowdsourcing experiment with 10 native and 10 non-native speakers in each language. Words that were highlighted by at least one crowdsourced annotator as difficult to understand in a given context (paragraph containing several sentences), were marked as complex in the final CWI-2018 datasets \citep{yimam-etal-2018-report}. The PorSimplesSent dataset, used for selecting sentences and target words for (Brazilian) Portuguese, is a corpus of original and manually simplified news articles. To identify complex words in the original sentences, the following procedure was used. An automatic word alignment tool was applied which marked inconsistencies between the original and simplified sentences. These were further checked by a native Brazilian-Portuguese speaker, who identified among them the complex words which contained simpler substitutes in the simplified sentences.

In all three datasets (English portion of CWI-2018, Spanish portion of CWI-2018, and PorSimplesSent for Portuguese), sentences often had several words marked as complex. For compiling the TSAR-2022 shared task datasets, we chose only one of the marked complex words as the target word, in each selected sentence. This made the task easier for participants, as they only had to take into account how the proposed simpler substitute fit the context (i.e., whether or not it preserves the original meaning) instead of additionally taking into account interactions among the proposed substitutes of different target words within the same sentence.

\subsubsection{Dataset Annotation}

To obtain a list of simpler substitutes for each target word, selected sentences (386 in English, 381 in Spanish, and 386 in Brazilian Portuguese) with marked target words were presented to crowdsourced workers who had a task of proposing a simpler substitute which would preserve the meaning of the original sentence. For English and Brazilian Portuguese, this crowdsourcing annotation task was done on Amazon Mechanical Turk,\footnote{\url{https://www.mturk.com/}} while for Spanish, it was done on Prolific platform.\footnote{\url{https://www.prolific.co/}} The annotation was first done for the Spanish dataset. The
guidelines used for the Spanish annotation were then translated
into English and Portuguese with minimal editing to ensure
that the task remained the same across languages. Details about dataset compilation and annotation across the three languages can be found in the work by \citet{StajnerEtAl-2022-FrontiersAI-LS}. Additional details about Spanish and Portuguese portions of the dataset can be found in the works by \citet{ferres-saggion:2022:LREC} and \citet{north2022alexsis}, respectively. The total number of annotated instances, the minimal, the maximal, and average number of proposed simpler substitutes per target word in each language are given in Table~\ref{tab:TSAR-Data}.

\begin{table}[]
    \begin{center}
    \begin{tabular}{l c c c r}
    \toprule
        \multirow{2}{*}{Language} & \multirow{2}{*}{Instances} & \multicolumn{3}{c}{Substitutes per target}\\ 
        & & Min & Max & Avg \\\midrule
        EN & 386 & 2 & 22 & 10.55 \\
        ES & 381 & 2 & 19 & 10.28 \\
        PT-BR & 386 & 1 & 16 & 8.10 \\
        \hline        
    \end{tabular}
    \caption{Statistics on the TSAR-ST 2022 multilingual lexical simplification dataset.}
    \label{tab:TSAR-Data}
\end{center}    
\end{table}

\subsubsection{Test Sets and Examples}

\begin{table}[]
    \begin{center}
    \begin{tabular}{l c r c}
    \toprule
        Language & Test & Trial & Total\\ 
        \midrule
        EN & 373 & 10 & 383\\
        ES & 368 & 12 & 380\\
        PT-BR & 374 & 10 & 384\\
        \hline        
    \end{tabular}
    \caption{Dataset splits for TSAR-2022 shared task. Instances with two or more repetitions of the complex word were excluded from the test set.}
    \label{tab:DataSplits}
\end{center}    
\end{table}

Annotated sentences in each language were split into trial and test datasets (Table~\ref{tab:DataSplits}).\footnote{Note that some instances had two repetitions of the complex word in the same sentence but were not included in the TSAR-2022 Shared Task splits of the Evaluation Benchmark. There was one such case in Spanish, three in English, and two in Portuguese.} Datasets are available under the Creative Commons Attribution-NonCommercial-ShareAlike 4.0 International License (CC-BY-NC-SA-4.0).\footnote{ \url{https://github.com/LaSTUS-TALN-UPF/TSAR-2022-Shared-Task/}} Examples of instances from trial  portion of the dataset are given in Table~\ref{tab:ExampleTrialTest}.

\begin{table*}[h]
    \begin{center}
    \begin{tabular}{l p{5cm} l p{7cm}}
    \toprule
        Data & Context/Sentence & Target & (Gold) Substitutes Ranked\\ 
        \midrule
        EN & A local witness said a separate group of attackers disguised in burqas — the head-to-toe robes worn by conservative Afghan women — then tried to storm the compound.	& disguised	& concealed:4, dressed:4, hidden:3, camouflaged:2, changed:2, covered:2, disguised:2, masked:2, unrecognizable:2, converted:1, impersonated:1\\
        \midrule
        ES & Floreció en la época clásica y tenía una reputada escuela de filosofía. &	reputada & 	prestigiosa:6, famosa:4, reconocida:2, afamada:2, conocida:2, renombrada:2, respetada:2, prestigioso:1, muy reconocida:1, valorada:1, acreditada:1, prestigiada:1\\
        \midrule
         
        PT & naquele país a ave é considerada uma praga	 & praga & 	peste:9, epidemia:5, maldição:3, doença:2, desgraça:2, tragédia:1, infestação:1\\
        \hline
    \end{tabular}
    \caption{Examples from the trial part of the TSAR 2022 dataset. The number after the ":" indicates the number of repetitions.}
    \label{tab:ExampleTrialTest}
\end{center}    
\end{table*}

\subsection{Baselines}

We provided two strong baselines: TSAR-TUNER and TSAR-LSBert.  TSAR-TUNER is an adaptation of the TUNER Lexical Simplification system \citep{ferresSG_BGNLP2017}, which is a state-of-the-art non-neural Spanish lexical simplification system. TSAR-TUNER differs from TUNER in that it omits the complex word identification and context adaptation phases. Instead, it returns an ordered list of substitution candidates. TSAR-TUNER sequentially executes four tasks: (1) sentence analysis, (2) word sense disambiguation, (3) synonyms ranking, and (4)
morphological generation. Details of TSAR-TUNER and its adaptation to English, and Portuguese can be found in \citep{StajnerEtAl-2022-FrontiersAI-LS}. 

TSAR-LSBert is an adaptation of LSBert \citep{qiang2020lsbert}, the state-of-the-art neural lexical simplification for English. LSBert uses the masked language model (MLM) of
BERT to predict a set of candidate substitution words and their
substitution probabilities. It combines five features to rank substitute candidates according to their simplicity: BERT prediction order, a BERT-based language model, the PPDB database, word frequency, and semantic word similarity from fastText word embeddings. Our TSAR-LSBert uses the same resources as the original system for lexical simplification in English. For lexical simplification in Spanish and Portuguese, all language-dependent components are adapted using the best available resources in corresponding languages. Details of TSAR-LSBert system and its adaptation to Spanish and Portuguese can be found in \citep{StajnerEtAl-2022-FrontiersAI-LS}.

\subsection{Evaluation Metrics}

To allow for fairer comparison of systems that propose a different number of substitution candidates, i.e.,\ not to penalize systems which return fewer candidates, all evaluation metrics are applied on a fixed number of $k$ top-ranked candidates. 

To account for various aspects of systems' performances, ten metrics were used as the official metrics of the shared task: ACC@1, MAP@k, potential@k,  accuracy@n@top1 where k $\in$ \{3, 5, 10\} and n $\in$ \{1, 2, 3\}.\footnote{ACC@1, 
MAP@1, and Potential@1 give the same results per definition. We thus used ACC@1 to denote them all in the official results.} 


{\bf Potential@k} is defined as the percentage of instances for which at least one of the $k$ top-ranked substitutes is also present in the gold data. 


{\bf Accuracy@k@top1} is defined as the percentage of instances where at least one of the $k$ top-ranked substitutes matches the most frequently suggested synonym in the gold data. Here is important to note that Accuracy@1@top1 was denoted as Accuracy@1 in \citep{StajnerEtAl-2022-FrontiersAI-LS}.

{\bf MAP@k}: The MAP metric is used commonly for evaluating information retrieval models and recommender systems 
\cite{Beitzel2018, Valcarce2020AssessingRM}. In the context of lexical simplification, instead of using a ranked list of relevant and irrelevant documents, we use a ranked list of generated substitutes, which can either be matched (relevant) or not matched (irrelevant) against the set of the gold-standard substitutes.
Unlike Precision@k, which only measures which percentage of the $k$ top-ranked substitutes can be found among the gold-standard substitutes, MAP@k additionally takes into account the position of the relevant substitutes among the first $k$ generated candidates (i.e.,\ whether or not the relevant candidates are at the top positions).




The evaluation script was provided to the participants and the research community.\footnote{\url{https://github.com/LaSTUS-TALN-UPF/TSAR-2022-Shared-Task}}

\section{Participating Systems}
\label{sec:systems}

We received the outputs of 13 teams for English, 6 for Spanish, and 5 for (Brazilian) Portuguese. Each team was allowed to submit outputs of up to 3 systems. This totaled to 31 submitted outputs for English, 15 for Spanish, and 14 for Portuguese. 

     \textbf{CILS} \citep{cils-tsar-2022-shared-task} submitted three systems for the English track. All systems use the Model Prediction Score and Embedding Similarity Score for candidate generation. A model prediction score is computed using the XLNet model \citep{yang_xlnet_2019} given the context and the target word with any word in the vocabulary of XLNet. The Embedding Similarity Score is the inner product of the embedding of the target word and the embedding of the respective word. The three systems differ in the ranking module. They rank the candidates based on different combinations of scores such as 1) the score from the candidate generation; 2) sentence similarity score (cosine similarity between the source and target sentence); 3) gloss sentence similarity score (the cosine similarity between the target word and the candidate); 4) WordNet score (a cosine similarity between the target word and the candidate extracted from WordNet); and 5) Validation score (a cosine similarity of the BERT-base between the source and target sentence).

    \textbf{PresiUniv} \citep{presiuniv-tsar-2022-shared-task} uses masked language model (follows LSBert) for candidate generation, ranks candidates by cosine similarity (extracted from FastText), and then filter them out by checking part-of-speech. The systems for the three languages are the same,  except that the language model is specific for each language. It is interesting that this approach works the best on Spanish dataset, but not so well on the Portuguese and English datasets (lower than the baseline).
    
    \textbf{UoM\&MMU} \citep{uom-mmu-tsar-2022-shared-task} uses an approach that consists of three steps: 1) candidate generation based on different prompt templates (e.g., <easier, simple> <word, synonym> for <target\_word>); 2) fine-tuning of a language model (BERT-based model) to select and rank candidates; and 3) post-processing to filter out noise and antonyms. This approach achieves the second rank on the Spanish dataset and the third rank on the English dataset, but to our surprise, the model ranks the lowest on the Portuguese dataset.

    \textbf{PolyU-CBS} \citep{polyu-cbs-tsar-2022-shared-task} proposes three approaches for the candidate ranking. In all three approaches, the candidates are generated using a masked language model. Then, the first approach ranks candidates based on the probability received from the candidate generation (base probability) and sentence probability extracted from GPT-2 pre-trained model by replacing the target word with its candidate. The second approach ranks candidates by base probability and masked language model scoring \citep{Salazar_2020}. The third model ranks candidates by base probability and contextualized embedding similarity (cosine similarity between the target word and its candidate in the context of the original sentence). Based on the official results, the third approach performs better than the other two in all languages.
    
    \textbf{CENTAL} \citep{cental-tsar-2022-shared-task} explored the use of masked language model for candidate generation with three strategies for context expansion: Copy, Query Expansion, and Paraphrase. The Copy strategy is a copy of the sentence itself (follows that of LSBert). The Query Expansion strategy extracts alternative words for the target word from FastText and then replaces the original sentence with each alternative word. The Paraphrase strategy (English only) extracts paraphrases from Pegasus \citep{zhang2020pegasus}. The authors propose three ranking approaches: 1) using the frequency of words generated by the three strategies; 2) training a binary classifier (English only) for the ranking; 3) the English ranking module (the binary classifier) performs cross-lingual ranking for Spanish and Portuguese. 
    
    \textbf{teamPN} \citep{teampn-tsar-2022-shared-task} proposes a model that extract candidates through a combination of modules such as verb sense disambiguation module (candidates are extracted from VerbNet \citep{schuler2005verbnet} and filtered by FitBERT \citep{havens2019fitbert}), paraphrase database module (PPDB) \citep{ganitkevitch2013ppdb}, DistilBERT module \citep{sanh2019distilbert} (uses masked language model), and Knowledge Graph module \citep{alberts-etal-2021-visualsem}.  Modules are combined depending on the part-of-speech of the target word. All extracted candidates are checked for correct inflection and ranked by FitBERT \citep{havens2019fitbert}.

    \textbf{MANTIS} \citep{mantis-tsar-2022-shared-task} adapts masked language model (RoBERTa) for candidate generation and performs the candidate ranking with three different approaches. The first ranking approach uses three features with different weights to rank the candidates: 1) pre-trained language model feature (the probability of the candidate extracted during the candidate generation), 2) Word Frequency, and 3) semantic similarity (cosine similarity between the FastText vector of the target word and the candidate). The second and third approaches rank candidates by word prevalence and equivalence score. The second approach uses crowd-sourcing word prevalence, which is a proportion of the population that knows a given word based on a crowd-sourcing study involving 220,000 people \citep{brysbaert2019word}. The third approach uses corpus-derived word prevalence, which is an estimate of the number of books that a word appears in \citep{johns2020estimating}. The equivalence score is the entailment score of the original sentence and the sentence replaced with the candidate. The experimental results have shown that the first approach performs better than the other two.

    \textbf{UniHD} \citep{unihd-tsar-2022-shared-task} submitted two systems. The first system was a zero-shot prompted GPT-3 with a prompt asking for simplified synonyms given a particular context. Simplifications are then  ranked. The second system was an ensemble over six different GPT-3 prompts/configurations with average rank aggregation. The second system attained the highest score for English on all metrics. The approach is simplistic in nature, relying heavily on the underlying language model which is only available for research through a paid interface.
    
    \textbf{RCML} \citep{rcml-tsar-2022-shared-task} proposes a system (English only) by applying the lexical substitution framework LexSubGen (based on XLNet) for candidate generation and ranks the candidates based on grammaticality (POS + morphological features), meaning preservation (BERTScore of the source and target sentences), and simplicity (predicted by an SVM classifier trained on CEFR level data). 
    
    \begin{table*}[ht]
        \begin{adjustbox}{width=1\textwidth,center=\textwidth}
    	\def\arraystretch{1.3}
    	\begin{tabular}{llll}
    		\specialrule{1.1pt}{1pt}{1pt}
    		\textbf{Team} & \textbf{Language} & \textbf{Approach} \\ \hline
    		CILS          & EN                & SG: Language Model (LM) probability and similarity score, SR: SG score, cosine similarity scores\\ 
    		PresiUniv     & EN, ES, PT        & SG: Masked Language Model (MLM), SR: cosine similarity, POS check     \\
    		UoM\&MMU      & EN, ES, PT        & SG: LM with prompt, SR: fined-tuned Bert model as classifier \\
    		PolyU-CBS     & EN, ES, PT        & SG: MLM, SR: MLM probability, GPT-2 probability, sentence probability, cosine similarity  \\
    		CENTAL       & EN, ES, PT        & SG: MLM, SR: word frequency, binary classifier\\
    		teamPN        & EN                & SG: MLM, VerbNet, PPDB, Knowledge Graph, SR: MLM probability  \\
    		MANTIS        & EN                & SG: MLM, SR: MLM probability, word frequency, cosine similarity \\
    		UniHD         & EN                & GTP-3 prompts: zero-shot, few-shot  \\
    		RCML          & EN                & SG: lexical substitution, SR: POS, BERTScore, SVM classifier \\
    		GMU-WLV       & EN, ES, PT        & SG: MLM, SR: MLM probability, word frequency \\
    		\specialrule{1.1pt}{1pt}{1pt}
    	\end{tabular}
    	\end{adjustbox}
    	\caption{The approaches taken by each team, categorised according to substitution generation (SG) and substitution ranking (SR) strategy.}
    	\label{table:approaches}
    \end{table*}

    \textbf{GMU-WLV} \citep{gmu-wlv-tsar-2022-shared-task} submitted two models for each of the three languages.   These two models follow the approach of LSBert, except the second model uses an additional Zipf frequency in the candidate ranking module.   The first model performs the best on the Portuguese dataset.

    A comparative  of system approaches is provided in Table \ref{table:approaches}.
    

\section{Results and Discussions}



In the following subsections we describe the results obtained by the participant teams for each of the tracks in the TSAR 2022 Multilingual Lexical Simplification shared task.\footnote{Please note that official results can also be queried at \url{https://taln.upf.edu/pages/tsar2022-st/\#results}}  Note that we will base our description on the ranking obtained by sorting submissions according to the ACC@1 metric as well as summarizing methods for which a paper has been submitted and accepted for the Shared Task (see Section \ref{sec:systems}). 

We also provided an extended version of the results,\footnote{\url{https://github.com/LaSTUS-TALN-UPF/TSAR-2022-Shared-Task/tree/main/results/extended}} which included ACC@1, Potential@k, MAP@k, macro-averaged Precision@k, macro-averaged Recall@k, and Accuracy@k@top1, for k $\in$ \{1, 2, ..., 10\}.
Precision@k and Recall @k were defined as follows:
\begin{itemize}

\item {\bf Precision@k}: the percentage of $k$ top-ranked substitutes that are present also in the gold data; 
 \item {\bf Recall@k}: the percentage of substitutions provided in the gold data that are included in the top $k$ generated substitutions.
\end{itemize}

\begin{table*}[ht]
\begin{adjustbox}{width=1\textwidth,center=\textwidth}
\begin{tabular}{lccccccccccc}
\specialrule{1.3pt}{1pt}{1pt}
Team &
  Run &
  \begin{tabular}[c]{@{}c@{}}ACC\\ @1\end{tabular} &
  \begin{tabular}[c]{@{}c@{}}ACC@1\\ @Top1\end{tabular} &
  \begin{tabular}[c]{@{}c@{}}ACC@2\\ @Top1\end{tabular} &
  \begin{tabular}[c]{@{}c@{}}ACC@3\\ @Top1\end{tabular} &
  \begin{tabular}[c]{@{}c@{}}MAP\\ @3\end{tabular} &
  \begin{tabular}[c]{@{}c@{}}MAP\\ @5\end{tabular} &
  \begin{tabular}[c]{@{}c@{}}MAP\\ @10\end{tabular} &
  \begin{tabular}[c]{@{}c@{}}Potential\\ @3\end{tabular} &
  \begin{tabular}[c]{@{}c@{}}Potential\\ @5\end{tabular} &
  \begin{tabular}[c]{@{}c@{}}Potential\\ @10\end{tabular} \\    \hline
UniHD &
  2 &
  \textbf{0.8096} &
  \textbf{0.4289} &
  \textbf{0.6112} &
  \textbf{0.6863} &
  \textbf{0.5834} &
  \textbf{0.4491} &
  \textbf{0.2812} &
  \textbf{0.9624} &
  \textbf{0.9812} &
  \textbf{0.9946} \\
UniHD           & 1 & 0.7721 & 0.4262 & 0.5335 & 0.5710 & 0.5090 & 0.3653 & 0.2092 & 0.8900 & 0.9302 & 0.9436 \\
MANTIS          & 1 & 0.6568 & 0.3190 & 0.4504 & 0.5388 & 0.4730 & 0.3599 & 0.2193 & 0.8766 & 0.9463 & 0.9785 \\
UoM\&MMU        & 1 & 0.6353 & 0.2895 & 0.4530 & 0.5308 & 0.4244 & 0.3173 & 0.1951 & 0.8739 & 0.9115 & 0.9490 \\
LSBert-baseline & 1 & 0.5978 & 0.3029 & 0.4450 & 0.5308 & 0.4079 & 0.2957 & 0.1755 & 0.8230 & 0.8766 & 0.9463 \\
RCML            & 2 & 0.5442 & 0.2359 & 0.3941 & 0.4664 & 0.3823 & 0.2961 & 0.1887 & 0.8310 & 0.8927 & 0.9436 \\
RCML            & 1 & 0.5415 & 0.2466 & 0.3887 & 0.4691 & 0.3716 & 0.2850 & 0.1799 & 0.8016 & 0.8847 & 0.9115 \\
GMU-WLV         & 1 & 0.5174 & 0.2493 & 0.3538 & 0.4477 & 0.3522 & 0.2626 & 0.1600 & 0.7533 & 0.8337 & 0.8981 \\
CL Lab PICT     & 1 & 0.5067 & 0.2064 & 0.3297 & 0.4021 & 0.3278 & 0.2331 & 0.1369 & 0.7265 & 0.7828 & 0.8042 \\
UoM\&MMU        & 3 & 0.4959 & 0.2439 & 0.3458 & 0.4235 & 0.3273 & 0.2411 & 0.1461 & 0.7560 & 0.8310 & 0.9088 \\
teamPN          & 2 & 0.4664 & 0.1823 & 0.3056 & 0.3378 & 0.2743 & 0.1950 & 0.0975 & 0.6729 & 0.7506 & 0.7506 \\
MANTIS          & 3 & 0.4611 & 0.2117 & 0.3351 & 0.4235 & 0.3227 & 0.2553 & 0.1673 & 0.7747 & 0.8793 & 0.9436 \\
teamPN          & 3 & 0.4504 & 0.1769 & 0.2841 & 0.3297 & 0.2676 & 0.1872 & 0.0936 & 0.6648 & 0.7399 & 0.7399 \\
teamPN          & 1 & 0.4477 & 0.1769 & 0.2815 & 0.3297 & 0.2666 & 0.1874 & 0.0937 & 0.6621 & 0.7453 & 0.7453 \\
PolyU-CBS       & 3 & 0.4316 & 0.2064 & 0.2788 & 0.3297 & 0.2683 & 0.1995 & 0.1178 & 0.6139 & 0.6997 & 0.7747 \\
MANTIS          & 2 & 0.4209 & 0.1662 & 0.2654 & 0.3565 & 0.2745 & 0.2193 & 0.1507 & 0.7131 & 0.8391 & 0.9517 \\
PresiUniv       & 1 & 0.4021 & 0.1581 & 0.2305 & 0.3002 & 0.2603 & 0.1932 & 0.1136 & 0.6568 & 0.7399 & 0.7962 \\
PolyU-CBS       & 1 & 0.3914 & 0.1823 & 0.2627 & 0.3002 & 0.2576 & 0.1883 & 0.1113 & 0.5924 & 0.6836 & 0.7533 \\
CILS            & 3 & 0.3860 & 0.1957 & 0.2627 & 0.3083 & 0.2603 & 0.2014 & 0.1267 & 0.5656 & 0.6005 & 0.6380 \\
CILS            & 2 & 0.3806 & 0.1903 & 0.2600 & 0.3083 & 0.2597 & 0.1997 & 0.1262 & 0.5630 & 0.6005 & 0.6434 \\
PresiUniv       & 3 & 0.3780 & 0.1474 & 0.2010 & 0.2573 & 0.2277 & 0.1609 & 0.0897 & 0.5656 & 0.6058 & 0.6327 \\
CILS            & 1 & 0.3753 & 0.2010 & 0.2788 & 0.3109 & 0.2555 & 0.1964 & 0.1235 & 0.5361 & 0.5898 & 0.6300 \\
CENTAL          & 2 & 0.3619 & 0.1152 & 0.2091 & 0.2788 & 0.2573 & 0.2056 & 0.1271 & 0.6541 & 0.7667 & 0.8418 \\
TUNER-baseline  & 1 & 0.3404 & 0.1420 & 0.1689 & 0.1823 & 0.1706 & 0.1087 & 0.0546 & 0.4343 & 0.4450 & 0.4450 \\
PolyU-CBS       & 2 & 0.3190 & 0.1447 & 0.2091 & 0.2573 & 0.1973 & 0.1490 & 0.0901 & 0.5120 & 0.6032 & 0.7104 \\
GMU-WLV         & 2 & 0.2815 & 0.0804 & 0.1689 & 0.2493 & 0.1899 & 0.1589 & 0.1200 & 0.5630 & 0.7399 & 0.8981 \\
CENTAL          & 1 & 0.2761 & 0.1313 & 0.1930 & 0.2117 & 0.1635 & 0.1183 & 0.0707 & 0.3780 & 0.4021 & 0.4182 \\
UoM\&MMU        & 2 & 0.2654 & 0.1367 & 0.2171 & 0.2680 & 0.1820 & 0.1307 & 0.0794 & 0.4906 & 0.5817 & 0.6756 \\
PresiUniv       & 2 & 0.2600 & 0.1018 & 0.1313 & 0.1554 & 0.1350 & 0.0862 & 0.0439 & 0.3136 & 0.3163 & 0.3163 \\
twinfalls       & 1 & 0.1957 & 0.0509 & 0.0884 & 0.1233 & 0.1175 & 0.0879 & 0.0535 & 0.3485 & 0.4235 & 0.5067 \\
twinfalls       & 2 & 0.1849 & 0.0643 & 0.0911 & 0.1367 & 0.1182 & 0.0857 & 0.0514 & 0.3565 & 0.4075 & 0.4664 \\
NU HLT          & 1 & 0.1447 & 0.0670 & 0.1018 & 0.1179 & 0.0902 & 0.0583 & 0.0301 & 0.2600 & 0.2815 & 0.2895 \\
twinfalls       & 3 & 0.0455 & 0.0107 & 0.0348 & 0.0455 & 0.0370 & 0.0277 & 0.0182 & 0.1474 & 0.2305 & 0.3619 \\
\specialrule{1.3pt}{1pt}{1pt}
\end{tabular}
\end{adjustbox}
\caption{Results submitted for the English track in comparison with the baselines (LSBert, TUNER). The best performances are in bold. Note: ACC@1, MAP@1, Potential@1, and Precision@1 give the same results as per their definitions}
\label{table:results_en}
\end{table*}

Overall, we observe that several systems achieved new state-of-the-art results  in the different tracks overtaking a previous competitive Neural Language Model for lexical simplification (LSBert).

\begin{table*}[ht]
\begin{adjustbox}{width=1\textwidth,center=\textwidth}
\begin{tabular}{lccccccccccc}
\specialrule{1.3pt}{1pt}{1pt}
Team & Run & \begin{tabular}[c]{@{}c@{}}ACC\\ @1\end{tabular} & \begin{tabular}[c]{@{}c@{}}ACC@1\\ @Top1\end{tabular} & \begin{tabular}[c]{@{}c@{}}ACC@2\\ @Top1\end{tabular} & \begin{tabular}[c]{@{}c@{}}ACC@3\\ @Top1\end{tabular} & \begin{tabular}[c]{@{}c@{}}MAP\\ @3\end{tabular} & \begin{tabular}[c]{@{}c@{}}MAP\\ @5\end{tabular} & \begin{tabular}[c]{@{}c@{}}MAP\\ @10\end{tabular} & \begin{tabular}[c]{@{}c@{}}Potential\\ @3\end{tabular} & \begin{tabular}[c]{@{}c@{}}Potential\\ @5\end{tabular} & \begin{tabular}[c]{@{}c@{}}Potential\\ @10\end{tabular} \\ \hline
GMU-WLV & 1 & \textbf{0.4812} & \textbf{0.2540} & \textbf{0.3716} & \textbf{0.3957} & \textbf{0.2816} & \textbf{0.1966} & \textbf{0.1153} & \textbf{0.6871} & \textbf{0.7566} & \textbf{0.8395} \\
CENTAL & 1 & 0.3689 & 0.1737 & 0.2433 & 0.2673 & 0.1983 & 0.1344 & 0.0766 & 0.5240 & 0.5641 & 0.6096 \\
PolyU-CBS & 3 & 0.3262 & 0.1390 & 0.1871 & 0.2139 & 0.1755 & 0.1256 & 0.0732 & 0.4491 & 0.5106 & 0.5748 \\
LSBert-baseline & 1 & 0.3262 & 0.1577 & 0.2326 & 0.2860 & 0.1904 & 0.1313 & 0.0775 & 0.4946 & 0.5802 & 0.6737 \\
PresiUniv & 1 & 0.3074 & 0.1604 & 0.2032 & 0.2379 & 0.1573 & 0.1077 & 0.0580 & 0.4598 & 0.5320 & 0.5935 \\
PresiUniv & 3 & 0.3048 & 0.1604 & 0.2032 & 0.2379 & 0.1555 & 0.1062 & 0.0571 & 0.4572 & 0.5294 & 0.5855 \\
PresiUniv & 2 & 0.2941 & 0.1604 & 0.1978 & 0.2326 & 0.1494 & 0.1020 & 0.0549 & 0.4411 & 0.5026 & 0.5588 \\
PolyU-CBS & 1 & 0.2807 & 0.1122 & 0.1470 & 0.1711 & 0.1515 & 0.1059 & 0.0629 & 0.3983 & 0.4705 & 0.5534 \\
CENTAL & 3 & 0.2245 & 0.0614 & 0.1310 & 0.1925 & 0.1478 & 0.1143 & 0.0769 & 0.4705 & 0.6096 & 0.8021 \\
PolyU-CBS & 2 & 0.2219 & 0.0882 & 0.1203 & 0.1497 & 0.1112 & 0.0797 & 0.0478 & 0.3315 & 0.3850 & 0.4919 \\
TUNER-baseline & 1 & 0.2219 & 0.1336 & 0.1604 & 0.1604 & 0.1005 & 0.0623 & 0.0311 & 0.2673 & 0.2673 & 0.2673 \\
GMU-WLV & 2 & 0.2165 & 0.0695 & 0.1363 & 0.2165 & 0.1559 & 0.1243 & 0.0845 & 0.5133 & 0.6550 & \textbf{0.8395} \\
CENTAL & 2 & 0.2058 & 0.0641 & 0.1203 & 0.1898 & 0.1470 & 0.1103 & 0.0726 & 0.4786 & 0.6016 & 0.7673 \\
UoM\&MMU & 1 & 0.1711 & 0.0695 & 0.0855 & 0.1096 & 0.1011 & 0.0747 & 0.0430 & 0.2486 & 0.2914 & 0.3636 \\
UoM\&MMU & 3 & 0.1577 & 0.0748 & 0.1016 & 0.1283 & 0.1071 & 0.0785 & 0.0461 & 0.2834 & 0.3262 & 0.4171 \\
UoM\&MMU & 2 & 0.1363 & 0.0454 & 0.0721 & 0.0962 & 0.0944 & 0.0711 & 0.0418 & 0.2379 & 0.2967 & 0.3609\\

\specialrule{1.3pt}{1pt}{1pt}
\end{tabular}
\end{adjustbox}
\caption{Results submitted for the Portuguese track in comparison with the baselines (LSBert, TUNER). The best performances are in bold. Note: ACC@1, MAP@1, Potential@1, and Precision@1 give the same results as per their definitions}
\label{table:results_pt}
\end{table*}

\subsection{English Track}
In table \ref{table:results_en} the results for English are presented sorted by ACC@1.\footnote{Note that the data to sort the results is available at \url{https://github.com/LaSTUS-TALN-UPF/TSAR-2022-Shared-Task/tree/main/results/official}} In this track, and of the 31 submitted runs, only four (from 3 teams: UniHD, MANTIS, and UoM\&MMU) performed better than the LSBert baseline according to ACC@1. Moreover, UniHD run number 2 achieved the best performance  in all the reported metrics. 
UniHD run number 2 outperforms the other teams' systems in more than 15 points in ACC@1 achieving a score of 0.8096 in this metric. This indicates that it is able to retrieve a correct synonym in the 80,96\% of instances of the dataset.
Moreover, UniHD's run number 2 achieves a 99,46\% of Potential@10. This indicates that it has the potential to retrieve at least one correct substitution in the top-10 predictions of almost all the instances.
In fact, it achieves 0.9624 in Potential@3 metric, which is almost nine points higher than the second best official result (MANTIS with 0.8900) and indicates also a great performance obtaining at least one correct substitution in the top-3 predictions.

It is important to highlight that the UniHD's system relies on a pre-trained {\em pay-per-query}  GPT-3 model to obtain candidate substitutions by prompting the  model with 6 versions of zero, one, and two shot prompts, based on the provided trial data,  finally combining the predicted candidate ranks to select the best substitutions. 
In contrast, team MANTIS relied on a freely available masked language model to obtain substitutions and an adaptation of the ranking procedure of LSBert.  While UoM\&MMU also relying on  freely available pre-trained masked language models fine tuned  with prompts to select the most appropriate substitutes and filtering substitution candidates by checking several resources (e.g. WordNet, corpora).  Also noticeable in the results of the task is that several ``neural'' systems under-perform the ``non-neural'' TUNER baseline in terms of ACC@1.  
Overall, it seems that the use of pre-trained masked language models fine-tuned to the task together with extra lexical resources or corpora produce very competitive approaches.

\subsection{Portuguese Track}
Table \ref{table:results_pt}, presents the results for Portuguese, also sorted by ACC@1.  In this track, and of the 14 submitted runs, only two (from two teams:  GMU-WLV and CENTAL) performed better than the LSBert baseline (as observed in the table, one team performed equally to LSBert in terms of ACC@1, but worst in the other metrics). The displayed results indicate that there is a clear top performing (considering all metrics)  system produced by team GMU-WLV. Surprisingly, they have relied on a simple approach to substitute generation and ranking by adopting a pre-trained Portuguese masked language model,  BERTimbau \cite{souza2020bertimbau}. The second best performing system according to ACC@1 is by the CENTAL team which also relied on a pre-trained masked language model in which several strategies were used to provide context to the target sentence, followed by a ranking procedure based on voting.  Similar as in the English track, in this track, several systems  under-perform, in terms of ACC@1,  the shared task non-neural TUNER baseline.

\subsection{Spanish Track}
Results for the Spanish track are presented in Table~\ref{table:results_es}. The systems' runs are sorted by ACC@1. Several systems outperformed the LSBert baseline. In particular, the PresiUniv team produced two competitive approaches which ranked first and third (tied with team UoM\&MMU). However, the approach did not reach top performance in several of the official metrics. The PresiUniv lexical simplifier relies on a masked language model approach for substitute selection combined with a word-embbeding similarity model for meaning  preservation and a filtering stage based on POS-tagging. The  UoM\&MMU, GMU-WLV and CENTAL (with approaches already described for English or Portuguese)  also performed well in the Spanish track, with UoM\&MMU and GMU-WLV achieving top scores for some of the metrics. The PolyU-CBS team produced a competitive system which  used a Spanish specific masked language model to generate substitutes and a ranking based on a combination of sentence language model probabilities and word-embedding similarities. Best performing systems in this track rely on Spanish-specific masked language models, corpus-based information, language model prompts, and  syntactic information, among others.


\begin{table*}[ht]
\begin{adjustbox}{width=1\textwidth,center=\textwidth}
\begin{tabular}{lccccccccccc}
\specialrule{1.3pt}{1pt}{1pt}
Team & Run & \begin{tabular}[c]{@{}c@{}}ACC\\ @1\end{tabular} & \begin{tabular}[c]{@{}c@{}}ACC@1\\ @Top1\end{tabular} & \begin{tabular}[c]{@{}c@{}}ACC@2\\ @Top1\end{tabular} & \begin{tabular}[c]{@{}c@{}}ACC@3\\ @Top1\end{tabular} & \begin{tabular}[c]{@{}c@{}}MAP\\ @3\end{tabular} & \begin{tabular}[c]{@{}c@{}}MAP\\ @5\end{tabular} & \begin{tabular}[c]{@{}c@{}}MAP\\ @10\end{tabular} & \begin{tabular}[c]{@{}c@{}}Potential\\ @3\end{tabular} & \begin{tabular}[c]{@{}c@{}}Potential\\ @5\end{tabular} & \begin{tabular}[c]{@{}c@{}}Potential\\ @10\end{tabular} \\ \hline
PresiUniv & 1 & \textbf{0.3695} & \textbf{0.2038} & \textbf{0.2771} & \textbf{0.3288} & 0.2145 & 0.1499 & 0.0832 & \textbf{0.5842} & 0.6467 & 0.7255 \\
UoM\&MMU & 3 & 0.3668 & 0.1603 & 0.2282 & 0.2690 & 0.2128 & 0.1506 & 0.0899 & 0.5326 & 0.6005 & 0.6929 \\
PresiUniv & 3 & 0.3614 & 0.2038 & 0.2581 & 0.2961 & 0.1944 & 0.1318 & 0.0706 & 0.5163 & 0.5543 & 0.5815 \\
UoM\&MMU & 2 & 0.3614 & 0.1603 & 0.2445 & 0.2907 & 0.2225 & 0.1657 & 0.0958 & 0.5380 & 0.6168 & 0.7010 \\
PolyU-CBS & 3 & 0.3586 & 0.1630 & 0.2010 & 0.2364 & 0.2068 & 0.1456 & 0.0850 & 0.5244 & 0.5978 & 0.6793 \\
GMU-WLV & 1 & 0.3532 & 0.1820 & 0.2635 & \textbf{0.3288} & 0.2202 & \textbf{0.1664} & \textbf{0.0994} & 0.5679 & \textbf{0.6793} & \textbf{0.7717} \\
UoM\&MMU & 1 & 0.3451 & 0.1494 & 0.2364 & 0.2907 & \textbf{0.2238} & 0.1614 & 0.0949 & 0.5543 & 0.6385 & 0.7038 \\
CENTAL & 1 & 0.3097 & 0.1467 & 0.2092 & 0.2391 & 0.1826 & 0.1327 & 0.0779 & 0.5000 & 0.5923 & 0.6358 \\
LSBert-baseline & 1 & 0.2880 & 0.0951 & 0.1440 & 0.1820 & 0.1868 & 0.1346 & 0.0795 & 0.4945 & 0.6114 & 0.7472 \\
PolyU-CBS & 1 & 0.2826 & 0.1141 & 0.1820 & 0.2255 & 0.1820 & 0.1320 & 0.0780 & 0.5000 & 0.5978 & 0.6820 \\
PresiUniv & 2 & 0.2500 & 0.1576 & 0.1793 & 0.1956 & 0.1197 & 0.0740 & 0.0371 & 0.3125 & 0.3152 & 0.3152 \\
GMU-WLV & 2 & 0.2364 & 0.0679 & 0.1304 & 0.1875 & 0.1557 & 0.1256 & 0.0833 & 0.4646 & 0.6168 & \textbf{0.7717} \\
CENTAL & 3 & 0.2201 & 0.0407 & 0.0896 & 0.1331 & 0.1416 & 0.1122 & 0.0745 & 0.4646 & 0.6086 & 0.7581 \\
PolyU-CBS & 2 & 0.2010 & 0.0869 & 0.1331 & 0.1739 & 0.1417 & 0.1025 & 0.0615 & 0.4103 & 0.4972 & 0.6413 \\
CENTAL & 2 & 0.1983 & 0.0652 & 0.1114 & 0.1657 & 0.1265 & 0.0979 & 0.0695 & 0.4184 & 0.5570 & 0.7282 \\
TUNER-baseline & 1 & 0.1195 & 0.0625 & 0.0788 & 0.0842 & 0.0575 & 0.0356 & 0.0184 & 0.1440 & 0.1467 & 0.1494 \\
OEG\_UPM & 1 & 0.1032 & 0.0434 & 0.0842 & 0.1086 & 0.0772 & 0.0594 & 0.0389 & 0.2527 & 0.3342 & 0.4456 \\

\specialrule{1.3pt}{1pt}{1pt}
\end{tabular}
\end{adjustbox}
\caption{Results submitted for the Spanish track in comparison with the baselines (LSBert, TUNER). The best performances are in bold. Note: ACC@1, MAP@1, Potential@1, and Precision@1 give the same results as per their definitions}
\label{table:results_es}
\end{table*}

\section{Conclusions and Further Work}

Lexical Simplification, the task of replacing difficult words in a sentence by easier to read or understand synonyms  preserving the meaning of the original sentence is an important problem which has gained considerable attention in the past few years.
In spite of its popularity for English, the task has attracted less research for other languages. Considering its social relevance in today's digital world, we put forward the first Shared Task on Multilingual Lexical Simplification addressing three languages: English, (Brazilian) Portuguese, and Spanish, and called the research community to challenge the state of the art. To carry out the task, we have prepared three datasets, one per each language, following similar data collection and data annotation approaches, leading the way to the development of future datasets for additional languages. The datasets are composed of sentences each containing a single complex word which needs to be simplified.  Although the datasets were intended only for testing the participating systems, a small (between 10 and 12 instances) portion was released as trial data. This was particularly useful for several teams to fine-tune their computational methods or prompts. The task also featured two baselines: one based on a competitive neural approach, and another one on a traditional (dictionary-based) pipe-lined architecture.
The Shared Task attracted a considerable number of participants with a total of 60 systems' runs submitted across the three tracks.  Several systems  outperformed the competition, setting a new benchmark in Lexical Simplification. It is observed that  pre-trained masked language models when fine-tuned to the lexical simplification task  produce very competitive approaches in combination with additional syntactic/lexical resources or corpora.



The submitted systems relied heavily on pre-trained language models, which are known to hallucinate (i.e., generate non-factual statements based on previously seen contexts). In the context of substitution generation, hallucination may indicate that incorrect simplifications are returned when the context is under-specified or unfamiliar. Further work to ensure that the simplifications generated by such systems are faithful to the original text and are factual in nature will help to engender a culture of security and trust in simplification research.

In our dataset, we have not considered a key aspect of simplification, which is the user. Our datasets assume that there is one correct simplification that is the best simplification for all users. In fact, our `best' simplification is collected from many users and is based on the most frequently returned simplification. It is interesting to note that when asked to simplify the same word in the same context, users will answer differently. It is logical to conclude then, that a simplification system must return a term which is appropriate to a user.

Concerning the selected evaluation metrics, although the MAP@K metric takes into account the order of returned items and it is very useful for cases when multiple relevant items are expected,
it has the disadvantage that the relevance of the returned items is binary. So, in further work, it could be included a metric that could take into account the possibility of graded or weighted relevance allowing the
participants to submit a weight associated to each prediction and allow ties.

Finally, we note that our evaluation methodology is entirely automated, due to the constraints of a shared task environment. Whilst this is very useful for developing systems for lexical simplification, we strongly encourage those working on in-production systems to directly evaluate the resulting systems with the user bases that they are intended for. Automated evaluation is secondary to human evaluation, and this is especially true in simplification where the goal is to enable the user to better understand the original information.

\section*{Acknowledgements}
We thank all the teams who registered and sent submissions to the Shared Task. We acknowledge partial support from the individual project
Context-aware Multilingual Text Simplification (ConMuTeS)
PID2019-109066GB-I00/AEI/10.13039/501100011033 awarded
by Ministerio de Ciencia, Innovación y Universidades (MCIU)
and by Agencia Estatal de Investigación (AEI) of Spain.

\bibliography{custom}
\bibliographystyle{acl_natbib}

\appendix



\end{document}